\documentclass{article}

\usepackage[main, final]{neurips_2026}
\makeatletter
\renewcommand{\@noticestring}{AI Agents for Discovery in the Wild (AID-Wild), Workshop at ACM CAIS 2026.}
\makeatother

\usepackage[utf8]{inputenc}
\usepackage[T1]{fontenc}
\usepackage{hyperref}
\usepackage{url}
\usepackage{booktabs}
\usepackage{amsfonts}
\usepackage{amsmath}
\usepackage{amssymb}
\usepackage{nicefrac}
\usepackage{microtype}
\usepackage{xcolor}
\usepackage{graphicx}
\usepackage{enumitem}
\usepackage{multirow}
\usepackage[table]{xcolor}
\usepackage{tikz}
\usepackage{graphicx}
\usetikzlibrary{arrows.meta, positioning, shapes.symbols, shapes.geometric, calc}

\usepackage{listings}
\usepackage{caption}

\newcommand{\benchname}{\textsc{StatMechBench-v0}}

\newcommand{\wanyu}[1]{}
\newcommand{\wynew}[1]{}
\newcommand{\todo}[1]{}
\newcommand{\wz}[1]{}

\title{Exploring Structures in Physics Problems: Can AI Agents Discover Statistical Mechanical Mappings?}

\author{
  Wanyu Zhao\thanks{Equal contribution.} \\
    Siebel School of Computing and Data Science\\
  University of Illinois Urbana-Champaign \\
  \texttt{wanyu2@illinois.edu}
  \And
  Wanbing Zhao\footnotemark[1] \\
  Department of Physics and Astronomy \\
  Rice University \\
  \texttt{wz56@rice.edu}
}

\begin{document}

\maketitle

\begin{abstract}



An important skill in theoretical physics is to recognize when a new problem can be transformed into a known model. We study this skill as an AI-agent task: can LLM-based agents discover statistical mechanical mappings from a raw partition function to a tractable representation? To probe this question, we introduce \benchname, a benchmark of six Ising-type problems covering transfer-matrix methods, gauge-removable disorder, and planar/Pfaffian structure. We evaluate a simple propose-verify-revise agent across multiple LLMs and problem phrasings. The results show that numerical feedback often helps agents repair code and recover correct partition functions. However, agents can also pass the numerical checks while misidentifying the underlying tractable class or understating computational complexity. This both reveals limitations in current LLM reasoning and calls for a verification stack that goes beyond numerical agreement, incorporating, for example, symbolic checks and structural invariants. Our study provides an early evaluation and design directions for AI agents aimed at structural discovery in theoretical physics.

\end{abstract}

\section{Introduction}




A common mindset in theoretical physics is to relate a new problem to existing theories, such as mapping newly formulated models to well-studied ones. Statistical mechanics, a key branch in physics, has developed a collection of paradigmatic models to understand how microscopic degrees of freedom give rise to macroscopic phases and critical phenomena~\citep{baxter1982exactly}. Examples include the Ising model~\citep{lenz1920beitrag,ising1925beitrag}, the Potts model~\citep{potts1952generalized}, and many other exactly solved lattice models~\citep{baxter1982exactly}. Developed and solved by generations of physicists over the past century, these models encode invaluable expert knowledge, and physicists seek transformations to connect new statistical mechanical problems to existing models, allowing known results, intuitions, and tools to be reused.

While successfully relating one statistical mechanical model to another can itself be a profound physical result, as exemplified by the Kramers--Wannier duality~\citep{kramers1941statisticsI}, it also serves as one step in a broader research workflow. For example, physicists identify a mapping from a new problem to an existing model and use it to compute quantities of interest, simplify the analysis, and gain qualitative understanding of the original problem. However, recognizing or constructing such mappings is nontrivial: it often requires familiarity with existing models, mathematical or physical intuition about the structure of the original problem, and the ability to apply suitable transformations. Historically, many important mappings have required substantial expert effort, and this difficulty can become a key bottleneck in research, especially for those outside the relevant subfield.

In this work, we ask: \textit{Can we build an LLM-based AI agent to discover statistical mechanical mappings?} Formally, we view a mapping as a transformation from the original problem representation to a reference model whose
solution structure is already known (\S\ref{sec:background}). Our key insight is that LLMs could have learned about all the existing physics models as they are described both mathematically and in natural language across textbooks, papers, and lecture notes, and an LLM-based agent may be able to propose plausible mapping hypotheses by combining model descriptions, symbolic patterns, and high-level physical analogies, with its reasoning and self-reflection abilities.


We first build our insight and understanding of current LLMs' ability in this domain through a systematic evaluation (\S\ref{sec:eval}). Specifically, we collect a set of representative statistical mechanical mapping problems and group them by difficulty into three tiers (\S\ref{subsec:three-tier}). Our preliminary evaluation on a subset of problems shows that LLMs can find the mappings and construct tractable partition-function representations correctly with feedback loops, under both canonical and paraphrased problem descriptions (\S\ref{subsec:exp_design}, \S\ref{subsec:exp_res}). We also identify several failure modes. These observations both reveal the ability of current AI and shed light on our next step of building a discovery agent. 

Based on our problem categories and evaluation findings, we propose future statistical mechanical discovery agent designs and challenges (\S\ref{sec:open}). Our ultimate goal is to support frontier research, e.g., error thresholds for quantum codes~\citep{dennis2002topological,chubb2021statistical} and phase-transition questions in monitored quantum dynamics~\citep{skinner2019measurement,bao2020theory}. More broadly, as an early attempt to concretely apply AI agents to open-ended theoretical physics problems, we hope this work encourages physicists to further explore the role of LLMs in research. By highlighting both the strengths and limitations of current agentic AI systems in assisting scientific discovery, this work aims to contribute to accelerating progress in science.

\section{Background and Problem Formulation}
\label{sec:background}





The aim of \textit{statistical mechanics} is to explain and predict macroscopic behavior of a system from its microscopic degrees of freedom and interactions. 

In this section, we provide necessary background on statistical mechanics, following Ref.~\citep{baxter1982exactly}, and then formalize the problem of ``discovering statistical mechanical mappings". We end the section with a review of recent progress in AI agents for scientific discovery that inspires this work.


\subsection{Partition function and exactly solvable models}

We begin with the partition function \(Z\), which allows macroscopic thermodynamic observables to be obtained from microscopic configurations and will serve as our starting point for constructing a statistical-mechanical mapping.


Classical equilibrium statistical mechanics is formulated in terms of the \textit{partition function}\footnote{For concreteness, we describe the canonical ensemble at fixed temperature. Other equilibrium ensembles have analogous partition functions.}:
\begin{equation}\label{eq:partition_func}
    Z = \sum_{s} e^{-\beta H(s)},
\end{equation}
where $s$ denotes a microscopic state, or configuration, of the system, e.g., for a system with $n$ binary degrees of freedom, $s = (s_1, \dots ,s_n)$ with $s_i \in \{-1,+1\} $ for $i\in\{1,\dots,n\}$; $H(s)$ denotes the energy of the configuration $s$; and $\beta$ is the inverse
temperature. \(Z\) is understood to depend on the physical model parameters entering $H(s)$ and on temperature; this dependence is left implicit here.
The sum---for discrete systems---becomes an integral and a trace for continuous and quantum systems respectively.
Partition function is central to statistical mechanics because normalizing \(e^{-\beta H(s)}\) by \(Z\) gives the probability distribution over microscopic configurations $s$ (i.e., \(p(s) = \frac{e^{-\beta H(s)}}{Z}\)); then, macroscopic observables can be obtained by taking weighted averages over microscopic states.





Although $Z$ links microscopic states to macroscopic observables, directly evaluating it as in Eq.~\ref{eq:partition_func} is intractable for realistic interacting\footnote{
Here, ``interacting'' means that the energy function  $H(s)$ contains terms involving more than one degree of freedom, for example, pairwise interactions between the $i$-th and $j$-th components of $s = (s_1,\dots,s_n)$.
} systems of macroscopic size, which contain an enormous number of degrees of freedom, typically on the order of Avogadro's number ($10^{23}$). For example, consider a system of $n$ binary degrees of freedom, i.e., $s=(s_1,\ldots,s_n)$ with
$s_i\in\{0,1\}$, a direct computation of Eq.~\ref{eq:partition_func} involves a sum over $2^n$ ($n \sim 10^{23}$)
configurations. Consequently, one or both of the following strategies is adopted: (i) replacing the real system by a simplified idealization (a \textit{model}), specified by the states $s$ and the energy function \(H(s)\) chosen to capture the relevant physics while giving the partition function a structure that makes it tractable. (ii) introducing approximations to evaluate the sum.

In special cases, the structure of the model allows the partition function to be evaluated exactly; such models are usually called \emph{exactly solvable models}.
To date, the catalog of exactly solvable models remains limited, but it forms a valuable theoretical library for physicists to use, marking the current boundary of analytical control. Table~\ref{tab:exact_models} summarizes four groups of models that have been exactly solved.






\usetikzlibrary{arrows.meta, positioning, shapes.symbols, shapes.geometric, calc}

\begin{figure}[!t]
\centering
\resizebox{0.65\linewidth}{!}{
\begin{tikzpicture}[
    node distance=8mm,
    >=Latex,
    box/.style={
        rectangle,
        rounded corners=2pt,
        draw=black,
        thick,
        align=center,
        minimum width=8.2cm,
        minimum height=1.05cm,
        font=\small,
        fill=gray!5
    },
    topbox/.style={
        rectangle,
        rounded corners=2pt,
        draw=black,
        thick,
        align=center,
        minimum width=8.2cm,
        minimum height=0.95cm,
        font=\small\ttfamily,
        fill=gray!10
    },
    arrow/.style={
        -{Latex[length=2.5mm]},
        thick
    },
    brain/.style={
        cloud,
        cloud puffs=14,
        cloud puff arc=120,
        draw=black,
        thick,
        fill=blue!6,
        align=center,
        aspect=2,
        minimum width=3.8cm,
        minimum height=2.0cm,
        font=\small
    }
]

\node[topbox] (model) {Model};

\node[box, below=of model] (map) 
{\textbf{I. Transformations}};

\node[box, below=of map] (solv) 
{\textbf{II. Solvability structures}};

\node[box, below=of solv] (eval) 
{\textbf{III. Evaluation}};

\node[topbox, below=of eval] (Z) {$Z$};

\draw[arrow] (model) -- (map);
\draw[arrow] (map) -- (solv);
\draw[arrow] (solv) -- (eval);
\draw[arrow] (eval) -- (Z);

\node[brain] (brain) at ($(solv.west)+(-3.6,0)$)
{Physicists'\\ reasoning};


\end{tikzpicture}
}
\caption{
General logic underlying the exact evaluation of a partition function \(Z\). 
}
\label{fig:workflow}
\end{figure}
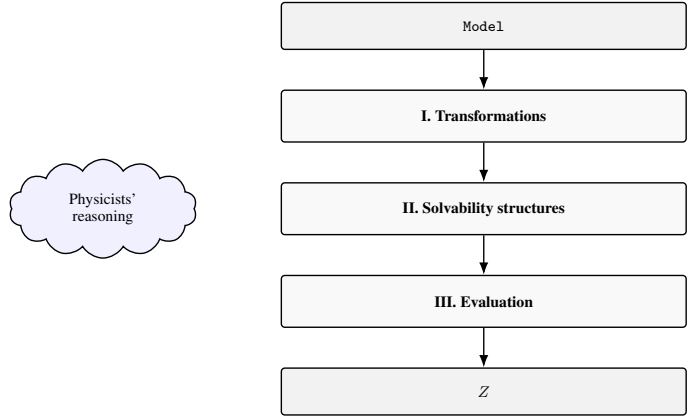

\subsection{Statistical mechanical mapping} 
\label{subsec:map}




We extract the general logic for physicists evaluating a partition function $Z$ in Figure~\ref{fig:workflow}: one maps a model to an equivalent formulation in which solvable structures become manifest, thereby enabling an explicit evaluation of the partition function. Similarly, this logic also guides the study of new models: when a model is first introduced to explain a new phenomenon, one natural first step in analyzing the model
is to seek transformations that reveal whether it has solvable structure or can be connected to a known model. 
To scale the search for transformations that may open tractable solution paths, we ask:


\emph{Can an LLM-based AI agent uncover the solvability structure of a problem by mapping it onto a known statistical mechanical model?}


 We refer to such a task as a \emph{statistical mechanical mapping}. As an initial probe, this work focuses on cases where the quantity of interest is presented as a raw partition function or a raw partition-function-type sum of the form in Eq.~\ref{eq:partition_func} (i.e., a sum over exponentially many configurations arising from the problem); accordingly, ``mapping'' means the mathematical transformations bringing the raw partition function into the partition function of a known model, e.g., changes of variables, gauge transformations, and duality
transformations. 
 

Generally, a successful partition-function mapping has both physical and methodological value. From a physical perspective, it can place a new problem within an existing
universality\footnote{In particular, if a model with the same dimensionality, symmetry, and interaction range as a given physical system can be solved, universality implies identical critical behavior, characterized by the same critical exponents near phase transitions.} class. From a methodological perspective, analytical control over the mapped model can be transferred
back to the original problem. Even when the mapped partition function is not exactly solvable, the mapping still provides physical insight and allows one to apply model-specific numerical methods, such as Monte Carlo or tensor network methods.

To the best of our knowledge, no prior work have investigated using LLM-based agent to identify statistical mechanical mappings. The closest work \citep{mlduality25gupta} uses machine learning algorithms to learn the resulting function mapped by duality (a specific transformation), while this work asks if an LLM agent can discover the mapping itself. 

\subsection{AI agents for scientific discovery}

Several LLM-based agents have been proposed for scientific and mathematical discovery through program search. FunSearch~\citep{funsearch24rp} introduced an LLM-guided evolutionary search framework that pairs a pretrained LLM with an automated evaluator to discover programs for mathematical and algorithmic problems. AlphaEvolve~\citep{alphaevolve} extends this line of work to a more general evolutionary coding agent that can modify larger codebases and optimize algorithms across scientific, mathematical, and engineering tasks. In theoretical physics, \cite{brenner2026solving} combine Gemini Deep Think, tree search, and automated numerical feedback to derive analytical solutions for a cosmic-string radiation problem by evaluating a core integral. It builds on a broader LLM--tree-search systems for scientific software generation called ERA~\citep{aisw2025aygun}, which formulate scientific software development as a scorable search problem and demonstrate results across multiple scientific domains. 

Several benchmarks have evaluated AI agents for physics problems. \cite{evalllmscience25song} evaluate LLMs on scenario-grounded scientific-discovery tasks across biology, chemistry, materials science, and physics (including solving an Ising model). PRL-Bench~\citep{prl-bench26miao} provides a benchmark for evaluating theoretical and computational physics research tasks derived from recent \textit{Physical Review Letters} papers, covering major subfields in modern physics including statistical physics.
This work targets the finer-grained problem of statistical mechanical mappings, where success requires identification of the underlying tractable structure, enabling a more precise diagnosis of whether an AI agent can discover a physically meaningful reduction rather than merely producing a plausible final answer or automating a broad research workflow.


\section{Probing LLM Agents for Statistical Mechanical Mapping}
\label{sec:eval}

\begin{table}[t]
\centering\small
\caption{Representative classes of exactly solved models, adapted from Ref.~\citep{baxter1982exactly}.}
\label{tab:exact_models}
\begin{tabular}{p{0.28\linewidth} p{0.68\linewidth}}
\toprule
\textbf{Class} & \textbf{Representative models and references} \\
\midrule
One-dimensional 
& Ising chain~\citep{ising1925beitrag}; 
one-dimensional Coulomb systems~\citep{lenard1961exact,baxter1963coulomb} \\

Infinite-dimensional 
& Mean-field and Bethe-lattice models; Kac-type long-range models 
(\cite{kac1963vanderwaalsI,uhlenbeck1963vanderwaalsII,hemmer1964vanderwaalsIII}) \\

Spherical model 
& Berlin--Kac spherical model and its infinite-spin-dimensional interpretation~\citep{montroll1949statistical,berlin1952spherical,stanley1968spherical} \\

Two-dimensional lattice 
& Ising model~\citep{kramers1941statisticsI,kramers1941statisticsII,onsager1944crystal}; 
ferroelectric/six-vertex models \cite{lieb1967fmodel,lieb1967kdp,lieb1967ice}; 
eight-vertex model~\citep{baxter1971eightvertex}; 
three-spin model~\citep{baxter1973threespin} \\
\bottomrule
\end{tabular}
\end{table}

As statistical mechanical mapping (\S\ref{subsec:map}) spans broad problem domains, touches diverse exactly solvable models and mathematical transformations, and can serve different purposes (e.g., computing quantity of interest vs. analyzing physics), and as today's LLM-based agent has a large design space (from prompt design to tool uses) with the LLM capability remaining opaque (e.g., memorization vs. reasoning), directly building a full end-to-end agent to achieve our task is unwieldy and challenging. Therefore, we conduct a probe evaluation to first build an understanding of LLM capabilities and guide the agent design. We first categorize statistical mechanical mapping problems into three tiers by difficulty and then evaluate a naive propose-verify-refine agent on a small benchmark comprising the easiest-tier problems (rediscover known ones). Our evaluation shows promising results of using feedback-guided LLM-based agents to identify mappings. We also reveal several failure modes that inform the agent design directions in \S\ref{sec:open}.

\subsection{Three-tier problems}

\label{subsec:three-tier}

We classify the statistical mechanical model mapping problems into three tiers by difficulty. We instantiate and evaluate a Tier 1 benchmark in \S\ref{subsec:exp_design}, with Tier 2 and 3 problem study left for future.

\paragraph{Tier 1: canonical models with known exact solution mechanisms.} Tier 1 consists of textbook canonical examples of exactly solved statistical mechanical models as listed in Table~\ref{tab:exact_models}. These models naturally compose a sanity-check benchmark because the mechanisms for evaluating their partition functions---e.g., transfer-matrix form, gauge-removable disorder, or planarity/Pfaffian structure---are well understood and analytically controlled. In this tier, the mapping agent's task is to recover a known exact solution/reduction or exact tractable representation given the model's original formulation.


\paragraph{Tier 2: established mappings in quantum information literature.}
    
  Tier 2 consists of known statistical mechanical mappings from the quantum information literature which are substantially more challenging to recover than those in Tier 1. Examples include toric-code threshold mappings with perfect or imperfect syndrome~\citep{dennis2002topological}, correlated-noise or subsystem-code generalizations~\citep{chubb2021statistical}, coherent-noise extensions~\citep{behrends2025statistical}, and replica mappings for monitored random circuits in the large local Hilbert-space dimension limit~\citep{bao2020theory}. In this tier, the mapping agent's task, given a raw partition function, is to recover an effective energy function $H(s)$ and identify any special parameter limits that enable the mapping. 
  
  

\paragraph{Tier 3: open but structured research problems.} Tier 3 consists of open research problems for which statistical mechanical reformulations are expected to be useful, but for which no sufficiently-tractable mapping is known. One example is maximum-likelihood threshold problems for structured quantum code families under various noise models, for which useful reductions remain unclear after a raw partition-function representation is obtained; another example is analytical approaches to measurement-induced entanglement phase transitions at finite local Hilbert-space dimension. In this tier, a useful agent output may be a partial reduction, an effective Hamiltonian, a candidate universality class, a numerically testable observable, or evidence that no simple mapping is likely.

\subsection{Experiment design}
\label{subsec:exp_design}

\begin{table}[th]
\centering\small
\caption{\benchname: six tasks, four tractable classes covered.
  $n$ is the number of spins; $L$ is the linear lattice size when applicable
  ($n=L^2$).}
\label{tab:tasks}
\begin{tabular}{llllll}
\toprule
\textbf{ID} & \textbf{Task} & \textbf{Probed $n$}  & \textbf{Tractable class} & \textbf{Complex.} \\
\midrule
P01 & 1D Ising chain, OBC, uniform $J$           & $\{6,10,14\}$ & \texttt{transfer\_matrix} & $O(n)$ \\
P02 & 1D Ising chain w/ random site fields, OBC  & $\{6,10,12\}$ & \texttt{transfer\_matrix} & $O(n)$ \\
P03 & 2D ferromagnetic Ising on $L\!\times\!L$ torus & $\{9,16\}$ & \texttt{transfer\_matrix} & $O(L\,2^{2L})$ \\
P04 & 2D Ising spin glass on planar grid  & $\{9,16\}$ & \texttt{pfaffian} & $O(n^3)$ \\
P05 & Mattis spin glass, rank-1 disorder         & $\{6,10,12\}$ & \texttt{gauge} & $O(n)$ \\
P06 & Random 3-regular $\pm 1$ Ising glass       & $\{8,10\}$    & \texttt{no\_efficient\_known} & --- \\
\bottomrule
\end{tabular}
\end{table}
\paragraph{Task problems.} We select six Tier 1 tasks, all classical Ising-type partition functions, listed in Table~\ref{tab:tasks}. We name this small benchmark \benchname. The tasks are chosen to span diverse mapping types: canonical transfer-matrix solvability (P01--P02), canonical gauge equivalence (P05), broadly-documented tractable reformulations via a two-dimensional transfer matrix
(P03), a Pfaffian reduction for zero-field planar Ising instances (P04), and a documented negative case with no assumed efficient exact mechanism in our predefined library (P06). 
Each task is labeled with a \textit{tractable class}\footnote{Tractable class is a unified name in operation level, it can be a target tractable model class, an algebraic transformation, an algorithm, or a graph structural property. Note that each task can be a multi-class problem because multiple tractability mechanisms could be applied. Here in \benchname, exactly one label under our consideration applies.} and the corresponding computational complexity. 


\paragraph{A naive propose-verify-refine mapping agent.}
\begin{itemize}[leftmargin=*]
    \item \textbf{Mapping proposer.} The proposer takes the natural-language task description and is asked to return a JSON object with fields \texttt{tractable\_class}, \texttt{complexity\_O}, \texttt{confidence}, \texttt{justification}, and \texttt{Z\_efficient(params, n)} (Python function) \footnote{We also do a simple integrity check of code at the Abstract Syntax Tree (AST) stage (e.g., illegal library imports).}. Task description includes the statistical mechanical mapping problem formulation with a list of tractable class labels (system prompt), the specific problem description (initial user prompt), and failure feedback from earlier runs (feedback user prompt). We make sure no tractable class label and corresponding complexity are leaked in the prompt. We provide nine tractable classes for each task to be classified into one. The full prompt template and tractable class list are provided in Appendix~\ref{app:prompts}. 
    \item \textbf{Brute-force numerical verifier.} We build a simple verifier that checks numerical equality by brute-force: a \texttt{Z\_brute(params, n)} function that returns the exact Z via summarization over all $2^n$ configurations.  Given a proposed \texttt{Z\_efficient(params, n)}, we launch an evaluation process with a timeout, write the candidate code into it, and pass a list of \texttt{(params, n)} probes. 
    A trial passes only if all probe results are within a tolerance (relative error $\varepsilon_{\text{rel}}=|Z_{\text{eff}}-Z_{\text{brute}}|/\max\bigl(|Z_{\text{brute}}|, 10^{-300}\bigr) < 10^{-6}$). On failure a brief textual report listing whether the code ran, the number and rate of failed probes, and an order-of-magnitude relative error (but not $Z_{brute}$ itself) is constructed as feedback of its round. We limit the number of rounds to $k$.
\end{itemize}


\subsection{Experiments and results}

\label{subsec:exp_res}



\paragraph{Setup and evaluation metrics.}
We cover eight model configurations across four vendors: Claude Haiku 4.5 \citep{anthropic_haiku45}, Claude Sonnet 4.6 \citep{anthropic_sonnet46}, Claude Opus 4.7 \citep{anthropic_opus47}, DeepSeek V4-Flash in Thinking mode \citep{deepseek_v4_flash,deepseek_thinking_mode}, DeepSeek V4-Pro \citep{deepseek_v4_pro}, Gemini 3.1-Pro \citep{google_gemini31pro}, and GPT-5.4-mini with minimal and high reasoning effort \citep{openai_gpt54mini}. We use $k=5$ verifier rounds, five parameter samples per probed $n$, and a per-trial timeout of 30\,s. Each task is associated with two problem descriptions: a canonical version and a paraphrased variant. Both descriptions are generated using LLMs and subsequently reviewed by a domain expert.

\begin{table}[t]
    \centering\small
    \caption{LLM performance on \benchname~across eight models with canonical and paraphrased problem descriptions. Opus 4.7 delivers the best overall performance.}
    \label{tab:results}
    \begin{tabular}{llccccrr}
    \toprule
    \textbf{Model} & \textbf{Variant} & \textbf{TC$_0$ (\%) $\uparrow$}
      & \textbf{NM$_0$ (\%) $\uparrow$}
      & \textbf{O (\%) $\uparrow$}
      & \textbf{$k^\star$ $\downarrow$}
      & \textbf{Time $\downarrow$} & \textbf{Cost $\downarrow$} \\
    \midrule
    \multirow{2}{*}{\textbf{\textsc{Haiku 4.5}}}
                  & canonical   & 100         &  50  & 100         & 4 &  1.6\,m & \$0.08 \\
                  & paraphrased     &  67         &  50  & 100         & 4 &  1.7\,m & \$0.09 \\
    \midrule
    \multirow{2}{*}{\textbf{\textsc{Sonnet 4.6}}}
                  & canonical   & 100                    &  83  & 83  & 1 &  1.7\,m & \$0.13 \\
                  & paraphrased & 83  & 100  & 83  & 0 &  1.6\,m & \$0.11 \\
    \midrule
    \multirow{2}{*}{\textbf{\textsc{Opus 4.7}}}
                  & canonical   & 100         &  83  & 100         & 1 &  1.7\,m & \$0.71 \\
                  & paraphrased & \textbf{100} & \textbf{100} & \textbf{100} & \textbf{0} &  1.2\,m & \$0.51 \\
    \midrule
    \multirow{2}{*}{\shortstack[l]{\textbf{\textsc{DeepSeek}}\\\textbf{\textsc{V4-Flash Think}}}}
                  & canonical   &  67         &  67  & 100         & 1 &  6.6\,m & \$0.07 \\
                  & paraphrased &  67         & 100  &  83         & 0 &  4.5\,m & \$0.05 \\
    \midrule
    \multirow{2}{*}{\shortstack[l]{\textbf{\textsc{DeepSeek}}\\\textbf{\textsc{V4-Pro}}}}
                  & canonical   & 100         &  83  & 100                          & 1 & 59.1\,m$^\dagger$ & \$0.26 \\
                  & paraphrased & 100 &  83 & 83$^*$    & 1 & 39.7\,m$^\dagger$ & \$0.19 \\
    \midrule
    \multirow{2}{*}{\textbf{\textsc{Gemini 3.1 Pro}}}
                  & canonical   &  83         &  83  &  83         & 2 & 16.0\,m & \$1.24 \\
                  & paraphrased & 83 &  83  &  83         & 2 & 18.8\,m & \$1.44 \\
    \midrule
    \multirow{2}{*}{\shortstack[l]{\textbf{\textsc{GPT-5.4-mini}}\\\textbf{\textsc{minimal}}}}
                  & canonical   & 100         & 100  &  83         & 0 &  0.5\,m & \$0.01 \\
                  & paraphrased &  67         & 100  &  67         & 0 &  0.6\,m & \$0.01 \\
    \midrule
    \multirow{2}{*}{\shortstack[l]{\textbf{\textsc{GPT-5.4-mini}}\\\textbf{\textsc{high}}}}
                  & canonical   &  83         &  83  & 100         & 3 & 18.4\,m & \$0.31 \\
                  & paraphrased & 83 &  83  & 100         & 2 & 12.6\,m & \$0.21 \\
    \bottomrule
    \end{tabular}

{\footnotesize $^\dagger$~DeepSeek-V4-Pro wall-clock includes connection-stall
overhead.\quad $^*$~False check: P03 emits
\texttt{complexity\_O =} O($n\,2^{\sqrt{n}}$), which
\emph{is} an honest acknowledgement of $2^L = 2^{\sqrt{n}}$, but the
audit's regex anchors on the literal letter \texttt{L}}
    \end{table}

\paragraph{How do different LLMs perform?} 
Table~\ref{tab:results} reports tractable class classification accuracy of the first round (TC$_0$), numerical evaluation pass rate of the first round (NM$_0$), post-hoc complexity honesty check score (O), saturation round number $k^\star$ where the agent succeeds with 100\% numerical evaluation pass and stops iteration, total wall-clock time and cost.
We omit numerical pass rate with the total $k\!=\!5$ loops from the table as all reach $100\%$, so on canonical Tier 1 tasks, the loop alone cannot discriminate frontier models from
each other. 

Overall, Opus 4.7 performs best, achieving the highest classification accuracy, numerical evaluation pass rate, and complexity honesty score, with low $k^*$. The results also show a clear improvement across Claude models from earlier to later generations and from DeepSeek V4-Flash Thinking to stronger DeepSeek V4-Pro. 

In several cases, models
pass the probed numerical tests while submitting exponential
row-transfer constructions under polynomial-time labels such as
\texttt{pfaffian}/$O(n^3)$. The O audit detects these complexity mismatches and lowers the corresponding complexity check scores ($<$ 100\%). Three models remain honest under O on both problem description variants: Haiku 4.5, Opus 4.7 and GPT-5.4-mini high. This also indicates that numerical correctness alone is insufficient to certify the claimed tractable class.

\paragraph{Does iterative refinement loop help?}

\begin{figure}[tb]
\centering
\includegraphics[width=0.9\textwidth]{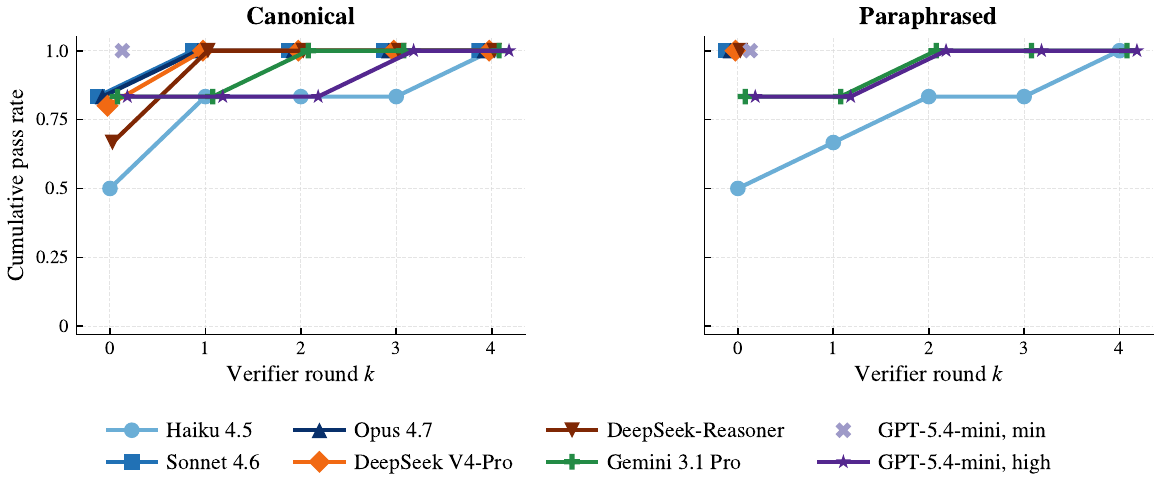}
\caption{Cumulative numerical evaluation pass rates across self-refinement rounds, with eight LLMs on canonical (left) and paraphrased (right) problem descriptions. A trial first passing
at round $j$ is counted as passed for all $k\geq j$. Most reasoning
models saturate by round 1, while only a few trajectories use much of
the $k=5$ budget. \wanyu{change marks}
}
\label{fig:pass_curve}
\end{figure}

Figure~\ref{fig:pass_curve} shows that numerical recovery usually occurs
quickly. Across models and problem description variants, many saturate within one verifier round.
Only a small number of trajectories use a
substantial fraction of the $k=5$ budget, typically when magnitude-only
feedback cannot localize the structural source of the error.

The recovery traces also show the loop's limitation. We find that when the initial
solution is numerically wrong, the verifier can often drive a successful
fallback, such as replacing a flawed Pfaffian attempt with a row-transfer
matrix. However, when the code is already numerically correct but carries
the wrong tractable-class label, the loop has little signal to correct
the label (e.g., DeepSeek V4-Flash Thinking and GPT-5.4-mini minimal on paraphrased description in Table~\ref{tab:results}). Thus the verifier is effective for numerical repair, but not
for detecting class-only mismatches.

\paragraph{How does it perform when we paraphrase the problem description?}

By looking at
$\Delta\mathrm{TC}_0 \equiv
\mathrm{TC}_{0,\text{paraphrased}}-\mathrm{TC}_{0,\text{canonical}}$, a finding is that reasoning-enabled LLMs are largely paraphrase-invariant, whereas low-effort or non-reasoning ones lose TC$_0$ accuracy under paraphrase. 

Another notable pattern is that paraphrasing can improve zero-shot numerical success (see NM$_0$ of Sonnet 4.6, Opus 4.7 and DeepSeek V4-Flash Thinking in Table~\ref{tab:results}). In P04, the canonical wording tends to trigger an ambitious Pfaffian construction,
which some models fail to implement reliably. The paraphrased wording
instead encourages a simpler row-transfer solution that passes at the
probed sizes; however, some still claim the \texttt{pfaffian} structure, yielding lower honesty. This reveals that LLMs may still rely on memorization rather than genuinely interpreting the mathematics and reasoning through the problem.
\paragraph{How did the agent go wrong?}

The evaluation reveals several failure modes. Sign errors produce large round-0 numerical mismatches
but are usually repairable by magnitude-only feedback; class-only errors can pass numerically while retaining the wrong tractable-class label, leaving the refinement loop with no signal for correction
and motivating richer verification checks such as symbolic checks. 
Canonical wording can induce memorized labels without reliable execution, which paraphrasing avoids by classifying to class with reliable execution but falsely claims the complexity, and some runs exhibit complexity false confidence: exponential
implementations are paired with polynomial-time claims and high self-confidence.
The negative control behaves as intended: models consistently identify
the no-known-efficient case and use brute force rather than forcing a spurious tractable label. Overall, these failures show that numerical verification alone is weak: it repairs many implementation bugs, but misses class-only and complexity-claim errors. 

\section{From v0 to Open Problems}
\label{sec:open}


The probing evaluation in \S\ref{sec:eval} demonstrates the initial promise of LLM-based agents for identifying statistical mechanical mappings, yet the benchmark is limited, and several failures occurred. Below, we outline how \benchname~can be systematically extended and discuss possible directions for improving the agent architecture.

\paragraph{Benchmark}
\begin{itemize}[leftmargin=*]
    \item \textbf{Task completion.} 
    \benchname~only covers a few tasks in Tier 1. Moving forward, the benchmark should cover a larger set of all three tiers' problems. Broader problems will expand the set of the ``tractable classes" as well, e.g., Tier 2 problems can require we do some algebraic transformation before having intuition about which tractable models they can potentially map to. Tier 3 open problems might not have mappings to existing physical models; thus, we may pivot the agent output to giving a bound instead, potentially by leveraging partial structure and approximate evaluation.
    \item \textbf{Task description.} We manually paraphrased the canonical problem description in \benchname. In later version of the benchmark, we could dynamically paraphrase with predefined templates.
    \item \textbf{Task scale.} We only cover $n\leq 16$. Solving practical problems needs to scale $n$, which will disable our brute-force numerical verifier.
\end{itemize}

\paragraph{Agent}

\begin{itemize}[leftmargin=*]

\item \textbf{Action space.}
One next step is to define a structured action library as the agent’s search space. 
Given a candidate partition function $Z$, an LLM--CAS (Computer Algebra System) pipeline first analyzes its structure by extracting the interaction hypergraph and computing invariants such as treewidth, planarity, symmetry group, and connectivity. 
These structural features determine a restricted set of admissible transformations, expressed in a domain-specific language (DSL). 
Within this constrained space, the agent performs symbolic manipulations---including gauge transformations, duality mappings, replica-limit evaluations, and Jordan--Wigner transformations---with each intermediate step checked for algebraic consistency. Correspondingly, the overall design will integrate multiple validation layers. The action library provides both hints and constraints, which could improve search efficiency yet should not overly constrain the space.

\item \textbf{Evaluation stack.}
A multi-layer evaluation stack combining static complexity analysis, symbolic verification via a CAS backend, and graph-theoretic invariants (e.g., treewidth, planarity, symmetry group, connectivity) can better support the mapping discovery. For ground-truth validation, we could extend brute-force enumeration with scalable numerical oracles, including MCMC sampling at the Nishimori temperature and tensor-network contraction on the Tanner graph, to enable system size of $n > 16$. Additional diagnostic signals, such as traceback-style checks, can be fed back into the agent context to improve robustness.

\item \textbf{Search and evolution.}
We could directly integrate with AlphaEvolve~\citep{alphaevolve} for evolutionary search or adopt a tree-search procedure following ERA~\citep{aisw2025aygun}. With tree search, each node corresponds to a candidate mapping procedure: the LLM proposes symbolic derivations, the CAS executes algebraic steps, and numerical oracles provide verifiable feedback. 
If the error $\varepsilon < \delta_{\mathrm{tol}}$, the candidate is promoted to larger system sizes and tested for consistency (e.g., phase-transition behavior). 
Otherwise, failed derivations or $\varepsilon > \delta_{\mathrm{tol}}$ incur penalties, and the search explores alternative hypotheses. 

\item \textbf{Prompting.} Stronger prompts could be designed and systematically explored. For example, negative prompting~\citep{brenner2026solving} could be used to encourage structural diversity across search branches and mitigate premature convergence. In addition, expert-provided hints could lead to more efficient search.

\item \textbf{Human oversight.}
Human-in-the-loop supervision can/should be incorporated to audit intermediate results, validate high-level reasoning, and guide exploration in ambiguous regimes.

\end{itemize}

\section{Conclusion}

This work takes a first step toward formulating statistical mechanical mapping as an agentic task for AI-assisted theoretical physics. The focus is distinguished from rigor-oriented tasks on exactly solvable models; here the agent task is to recover representations whose usefulness includes their physical relevance to the original problem, rather than by formal equivalence alone.  

We focus on a restricted setting---raw partition-function expressions with known tractable structure---and ask whether LLM-based agents can recover useful representations from such inputs. To this end, we introduce a three-tier taxonomy of statistical-mechanical mapping problems and instantiate its first tier---canonical models with known exact solution mechanisms---in \benchname. Although small in scale, \benchname\ provides a controlled testbed for studying the capabilities of LLM-based agents on this class of problems. Our experiments show that verifier feedback can help agents repair implementation errors and reproduce finite-size partition functions. At the same time, our failure studies reveal limitations in current LLM reasoning and point toward future mapping-discovery agents equipped with a multi-layer evaluation stack, a structured action space, and more effective evolutionary or search mechanisms.

\bibliographystyle{plainnat}
\bibliography{references}

\clearpage

\appendix
\section*{Appendix}
The appendices document everything needed to inspect the agent's
behavior and replicate the probe evaluation. App.~\ref{app:prompts} reproduces the
proposer prompts verbatim. App.~\ref{app:tasks} gives one card per task in
\benchname, with both the canonical and paraphrased descriptions.
App.~\ref{app:verifier} documents the sandboxed verifier
(AST allowlist, subprocess construction, feedback string template).



\section{Proposer prompts}
\label{app:prompts}
The proposer is steered by a single system prompt plus one of two user
templates (initial vs.\ feedback round).

\paragraph{System prompt.} The system prompt string:
\begin{lstlisting}
You are a research assistant participating in a benchmark of partition-function
mapping discovery.

Your job: given a description of a statistical-mechanics model H(s), produce
a JSON object that classifies it into a known tractable class and provides
an EXECUTABLE Python function `Z_efficient(params, n)` that computes the
partition function

    Z = sum_{s} exp(-beta * H(s))

ideally in time polynomial in n (or, where the problem is intractable in
the worst case, an honest brute-force fallback).

{LIBRARY}

Output FORMAT (a single JSON object; a Markdown ```json fence is fine but not required):

{
  "tractable_class": "<one of the library labels above>",
  "complexity_O": "<e.g. 'O(n)' or 'O(n^3)' or 'O(2^n)'>",
  "confidence": <a float in [0, 1]>,
  "justification": "<2-6 sentences: which transformation makes Z tractable and why>",
  "code": "<pure Python source defining Z_efficient(params, n) -> float; only standard libs + numpy>"
}

Rules:
  - The code MUST define a top-level function `Z_efficient(params, n)` returning a Python float (or any value losslessly convertible to one).
  - You may import `math`, `numpy`, `scipy`, `itertools`, `functools`. Do NOT import `os`, `sys`, `subprocess`, `socket`, or any networking/IO module, and do not read or write files.
  - If you genuinely believe no polynomial algorithm exists, set `tractable_class` to `"no_efficient_known"` and provide an honest brute-force implementation; do NOT pretend.
  - Your code will be executed in a subprocess with a 30 s timeout; budget your enumeration accordingly.
  - Your code will be checked numerically against brute-force enumeration with rtol=1e-6.
\end{lstlisting}

\paragraph{Tractable-class library (\texttt{LIBRARY}).}
Substituted into the system prompt. The labels here are the only legal
values for \texttt{tractable\_class}.
\begin{lstlisting}
Library of tractable partition-function classes (you must classify into one):

  * "transfer_matrix" -- 1D / quasi-1D systems; row-by-row update; O(n) or O(n * 2^L).
  * "free_fermion" -- quadratic fermionic Hamiltonian after Jordan-Wigner; O(n^3) via diagonalization.
  * "pfaffian" -- planar Ising / dimer model; Kasteleyn orientation; O(n^3).
  * "gauge" -- disorder removable by a sign-flip gauge (e.g. Mattis); reduces to known model.
  * "percolation" -- bond/site percolation universality; analytic critical points.
  * "rbim_nishimori" -- random-bond Ising on the Nishimori line (e.g. toric code threshold).
  * "treewidth_bounded" -- junction-tree algorithm; O(2^k * n) for treewidth k.
  * "exactly_solvable_other" -- ANY OTHER known polynomial-time class; you must name it explicitly.
  * "no_efficient_known" -- no known polynomial-time algorithm; honest brute force is acceptable.
\end{lstlisting}

The tractable class library mixes categories at different operational levels; Table~\ref{tab:class-details} further explains the underlying types. 
\begin{table}[h]
\centering\small
\caption{Details of tractable classes selected in this work.}
\label{tab:class-details}
\begin{tabular}{lp{0.65\textwidth}}
\toprule
Label & What kind of thing it actually is \\
\midrule
\texttt{transfer\_matrix}        & \texttt{Algorithm}---works on any 1D / quasi-1D Hamiltonian \\
\texttt{pfaffian}                & \texttt{Algorithm} (Kasteleyn)---applicable when the graph is planar \\
\texttt{treewidth\_bounded}      & \texttt{Graph property}---enables the junction-tree algorithm \\
\texttt{free\_fermion}           & \texttt{Representation}---quadratic-fermion form reachable by Jordan--Wigner; algorithm is diagonalisation \\
\texttt{gauge}                   & \texttt{Algebraic transformation}---sign-flip gauge that reduces to a known solvable model \\
\texttt{rbim\_nishimori}         & \texttt{Model + special line}---RBIM is generally hard; only on the Nishimori line is $Z$ analytically pinned \\
\texttt{percolation}             & \texttt{Universality / mapping}---Fortuin--Kasteleyn-style; gives analytic critical points but not a poly-time $Z$ in general \\
\texttt{exactly\_solvable\_other}& Escape hatch for any other known poly-time class (Bethe ansatz, Lieb's ice, \ldots) \\
\texttt{no\_efficient\_known}    & \texttt{Negative} label---honest brute force only \\
\bottomrule
\end{tabular}
\end{table}

\paragraph{Initial user prompt.} Round 0 of every trial.
\begin{lstlisting}
Here is the problem.

{task_description}

Please respond with ONLY the JSON object described in the system instructions.
\end{lstlisting}

\paragraph{Feedback user prompt.} Rounds $k\ge 1$ of every trial. The \texttt{\{prev\_json\}} slot is filled with the proposer's raw round-$(k-1)$ response; \texttt{\{feedback\}} is the verifier string from App.~\ref{app:verifier}.
\begin{lstlisting}
You previously responded to this problem:

--- problem ---
{task_description}
--- end problem ---

Your previous response was:
{prev_json}

The verifier reports:
{feedback}

Please revise. Reply with a NEW JSON object only (same schema). If you can
identify the bug in your previous code, fix it. If your tractable class was
wrong, choose a different one.
\end{lstlisting}

\section{Task cards (StatMechBench-v0)}
\label{app:tasks}
One card per task. Each card shows the tractable class and complexity, the probed sizes $n$, the canonical description, the paraphrased description, and the optional hint (unused in the evaluation).

\paragraph{How the paraphrases were produced.}
We followed a fixed rewrite rule to generate the paraphrased description:
\begin{enumerate}[leftmargin=12pt]\itemsep0pt
\item \textbf{Strip canonical model names.} ``Ising'', ``spin glass'', ``Pfaffian'', ``Mattis'', ``Hamiltonian'', ``ferromagnet'', ``torus'' are removed; geometric structure is described in plain words (``$L\times L$ grid of binary variables that wraps in both directions'' in place of ``$L\times L$ Ising torus'').
\item \textbf{Rename the spin variable.} The canonical statement uses $s_i\in\{-1,+1\}$; the paraphrase uses $x_i$ to remove the near-deterministic ``spin'' association.
\item \textbf{Preserve the $Z$ expression literally.} The number of configurations ($2^n$), the index ranges, the coupling structure, and the exponent inside $\exp(\cdot)$ are reproduced unchanged. P05 is the strongest illustration: the canonical statement names the Mattis model
and gives the rank-1 form $J_{ij}=\xi_i\xi_j$; the paraphrase
substitutes the already-contracted scalar $M(x)=\sum_i\xi_i x_i$ and asks for $\sum_x \exp\!\bigl((\beta/2)(M(x)^2-n)\bigr)$ without naming the model.
\item \textbf{Do not add or remove information.} The paraphrase contains no extra hint and no extra constraint relative to the canonical description; the optional hint field is the only place hints live.
\end{enumerate}
The fixed-string design makes the paraphrase set a reproducible benchmark artifact. Its quality is bounded by the curation rules; we view this as acceptable for v0 and discuss \emph{programmatic} rewriting (templated variable renaming, structural isomorphism over a graph DSL) as a v1 extension in \S\ref{sec:open}.

\paragraph{P01---1D Ising chain, OBC.} Tractable class \texttt{transfer\_matrix}, gold complexity $O(n)$. Probed sizes $n\in\{6,10,14\}$.
\begin{lstlisting}
[BLIND]
You are given the **1D Ising chain with open boundary conditions**.
Hamiltonian:  H(s) = -J * sum_{i=0}^{n-2} s_i * s_{i+1}, with s_i in {-1,+1}.
Goal: write a Python function
    def Z_efficient(params: dict, n: int) -> float:
        # params has keys "beta" (float) and "J" (float)
that computes the partition function
    Z(beta, J, n) = sum_{s in {-1,+1}^n} exp(-beta * H(s))
in time polynomial in n (NOT exponential).

[PARAPHRASED]
You are given a sequence of n binary variables x_1, ..., x_n with each
x_i in {-1, +1}, plus two real parameters beta and J. Compute
    Z(beta, J, n) = sum over all 2^n binary tuples (x_1,...,x_n) of
                       exp( beta * J * sum_{i=1}^{n-1} x_i * x_{i+1} ).
Naive enumeration of all 2^n tuples is too slow; your implementation should
run in time polynomial in n.
\end{lstlisting}

\paragraph{P02---1D Ising chain with random site fields, OBC.} Tractable class \texttt{transfer\_matrix}, $O(n)$.
Probed $n\in\{6,10,12\}$.
\begin{lstlisting}
[BLIND]
You are given a **1D Ising chain with random site fields and open boundary conditions**.
Hamiltonian:  H(s) = -J * sum_{i=0}^{n-2} s_i s_{i+1} - sum_{i=0}^{n-1} h_i s_i.
Implement
    def Z_efficient(params: dict, n: int) -> float:
        # params has "beta" (float), "J" (float), "h" (list of length n)

[PARAPHRASED]
You are given a sequence of n binary variables x_1, ..., x_n in {-1, +1},
a coupling constant J, and an array of n per-position weights h_1, ..., h_n.
Compute
    Z(beta, J, h, n) = sum over all 2^n tuples x of
        exp( beta * J * sum_{i=1}^{n-1} x_i * x_{i+1}
            + beta * sum_{i=1}^{n} h_i * x_i ).
\end{lstlisting}

\paragraph{P03---2D ferromagnetic Ising on $L\!\times\!L$ torus.} Tractable class \texttt{transfer\_matrix}, gold complexity $\text{poly}(L)\cdot 2^{L}$.
Probed $n\in\{9,16\}$ (i.e.\ $L\in\{3,4\}$).
\begin{lstlisting}
[BLIND]
You are given the **2D ferromagnetic Ising model on an L x L torus**
(periodic boundary conditions in both directions).
H(s) = -J * sum_{<i,j>} s_i s_j over all nearest-neighbour pairs.
Z must be computed in time polynomial in L (so exponential in sqrt(n) is
acceptable, but exponential in n is not).

[PARAPHRASED]
L x L grid of binary variables x_{r,c} in {-1,+1}; the grid wraps in both
directions. E is the set of all unique neighbour pairs (2*L^2 entries).
Compute Z(beta, J, n) = sum over 2^n assignments of
       exp( beta * J * sum_{(i,j) in E} x_i * x_j ).
[HINT, unused] Consider a row-to-row transfer matrix of size 2^L by 2^L.
\end{lstlisting}

\paragraph{P04---2D Ising spin glass on planar grid (open).} Tractable class \texttt{pfaffian}, $O(n^3)$. Probed $n\in\{9,16\}$.
\begin{lstlisting}
[BLIND]
You are given a **2D Ising spin glass on an L x L OPEN-boundary planar
lattice** with random binary couplings on each edge.
H(s) = - sum_{horiz edges} Jh[r,c] * s[r,c] * s[r,c+1]
       - sum_{vert  edges} Jv[r,c] * s[r,c] * s[r+1,c].
Z must be polynomial in n (NOT exponential).

[PARAPHRASED]
L x L grid of binary variables x_{r,c}; grid does NOT wrap. Each horizontal
edge carries a binary weight Jh[r][c] in {-1,+1}; each vertical edge
Jv[r][c] in {-1,+1}. Compute
   Z(beta, Jh, Jv, n) = sum over all 2^n assignments x of
      exp( beta * sum_{r,c: c<L-1} Jh[r][c] * x_{r,c} * x_{r,c+1}
         + beta * sum_{r,c: r<L-1} Jv[r][c] * x_{r,c} * x_{r+1,c} )
[HINT, unused] The interaction graph is planar.
\end{lstlisting}

\paragraph{P05---Mattis spin glass (rank-1 disorder).} Tractable class \texttt{gauge}, $O(n)$. Probed $n\in\{6,10,12\}$.
\begin{lstlisting}
[BLIND]
**Mattis-like spin glass on n spins** with all-to-all coupling.
J_{ij} = xi_i * xi_j for i != j, with xi a fixed +/- 1 pattern.
H(s) = -(1/2) * sum_{i!=j} J_{ij} s_i s_j = -(1/2) * [ (sum_i xi_i s_i)^2 - n ].
Z(beta, xi, n) in time polynomial in n.

[PARAPHRASED]
n binary variables x_1,...,x_n in {-1,+1} plus a fixed pattern vector xi.
For each x define M(x) = sum_i xi_i * x_i. Compute
   Z(beta, xi, n) = sum over 2^n assignments of exp( (beta/2) * (M(x)^2 - n) ).
[HINT, unused] The coupling matrix J_{ij} = xi_i * xi_j is rank-1.
\end{lstlisting}

\paragraph{P06---Random 3-regular $\pm 1$ Ising glass.}
Tractable class \texttt{no\_efficient\_known}, $O(2^n)$ (best known on general instances). Probed $n\in\{8,10\}$.
\begin{lstlisting}
[BLIND]
**Ising spin glass on a random 3-regular graph** with binary +/- 1 couplings.
Edges given as edge list; H(s) = -sum_k J[k] * s[edges[k][0]] * s[edges[k][1]].
"If you determine, after analysing the structure of the interaction graph,
that no polynomial-time algorithm in any class from the library applies,
state so explicitly in your justification and provide your best
implementation; the verifier will only call you at small n."

[PARAPHRASED]
Graph on n vertices specified by edge list, every vertex has degree 3.
Compute Z(beta, edges, J, n) = sum over 2^n assignments of
  exp( beta * sum_k J[k] * x[edges[k][0]] * x[edges[k][1]] ).
\end{lstlisting}

\section{Verifier internals}
\label{app:verifier}
The verifier is an integrity check on a cooperative proposer, not an adversarial sandbox. Its job is to (i)
reject obvious filesystem / network access at parse time, (ii) execute the candidate \texttt{Z\_efficient} on a list of $(n,\text{params})$ probes in isolation, and (iii) compare each returned value against $Z_{\text{brute}}$ at $\varepsilon_{\text{rel}}=10^{-6}$.

\paragraph{AST import allowlist.}
A static import check walks the candidate's AST and rejects any \texttt{import} or \texttt{from} statement whose top-level module is not in:
\begin{lstlisting}
ALLOWED_IMPORTS = {
  "math", "numpy", "scipy", "itertools", "functools",
  "operator", "fractions", "collections", "cmath", "decimal",
  "dataclasses", "typing",
}
\end{lstlisting}
Sub-modules thereof (e.g.\ \texttt{scipy.linalg}) are allowed.
\texttt{os}, \texttt{sys}, \texttt{subprocess}, \texttt{socket}, \texttt{pathlib}, \texttt{requests}, \ldots\ are all rejected before the subprocess starts. The allowlist only exists to surface accidental IO, so any adversarial bypasses (\texttt{\_\_import\_\_(`os')}, monkey-patched \texttt{getattr} chains) may still pass.

\paragraph{Subprocess template.}
Every probe set is run in a freshly-launched Python subprocess via \texttt{subprocess.run} with a 30s wall-clock timeout, \texttt{stdin} carrying a JSON list of calls and \texttt{stdout} carrying a single sentinel-prefixed line:
\begin{lstlisting}
import json, sys, math, itertools, functools
import numpy as np

def _to_float(x):
    try: return float(x)
    except (TypeError, ValueError):
        arr = np.asarray(x)
        if arr.size == 1: return float(arr.reshape(()).item())
        raise TypeError(f"Z_efficient returned non-scalar of shape {arr.shape}")

# ---------- candidate code starts ----------
__CANDIDATE_CODE__
# ---------- candidate code ends ----------

if 'Z_efficient' not in dir():
    print("__VERIFIER_RESULT__" + json.dumps({"error": "missing Z_efficient"}))
    sys.exit(0)

req = json.loads(sys.stdin.read())
out = []
for call in req["calls"]:
    try:
        z = Z_efficient(call["params"], call["n"])
        out.append({"ok": True, "z": _to_float(z)})
    except Exception as e:
        out.append({"ok": False, "error": f"{type(e).__name__}: {e}"})
print("__VERIFIER_RESULT__" + json.dumps({"results": out}))
\end{lstlisting}
The sentinel prefix \texttt{\_\_VERIFIER\_RESULT\_\_} isolates the result line from any debug \texttt{print} the candidate may emit; The subprocess \texttt{stdout} parser scans the lines in reverse and returns the first line it encounters with this prefix.

\paragraph{Environment-variable allowlist.}
The subprocess inherits only:
\begin{lstlisting}
PATH, HOME, LANG, LC_ALL, LC_CTYPE,
TMPDIR, TEMP, TMP, PYTHONPATH,
LD_LIBRARY_PATH, DYLD_LIBRARY_PATH.
\end{lstlisting}
Every other environment variable is dropped, so API keys and other credentials in the parent shell are not exposed to candidate code.

\paragraph{Probe construction.}
For each $n$, we
draw some independent parameter samples (default 5 samples) with a random seed, so the probe set is deterministic given \texttt{(seed, task)}. $Z_{\text{brute}}$ is computed in the parent process; only the candidate's $Z_{\text{eff}}$ runs in the subprocess.

\paragraph{Pass criterion.} A trial \emph{passes} only if every probe returned a finite float and
satisfied $\varepsilon_{\text{rel}}=|Z_{\text{eff}}-Z_{\text{brute}}|/\max\bigl(|Z_{\text{brute}}|, 10^{-300}\bigr) < 10^{-6}$.

\paragraph{Feedback string template.}
The verifier also produces a \texttt{\{feedback\}} string, which is pasted into the next round's user prompt (see App.~\ref{app:prompts}). Three branches:
\begin{lstlisting}
# (a) Code didn't run at all:
"Your code did not run. Error: {report.error}"

# (b) Code ran but every probe agreed:
"Verifier ran on {N} (n, params) samples.
All samples passed (max relative error {max_rel_err:.2e})."

# (c) Code ran and at least one probe disagreed:
"Verifier ran on {N} (n, params) samples.
{k}/{N} samples failed.
  - relative error {rel_err:.2e} >> 1e-06 -- your computed Z disagrees with brute force.
  - relative error {rel_err:.2e} >> 1e-06 -- your computed Z disagrees with brute force.
  - relative error {rel_err:.2e} >> 1e-06 -- your computed Z disagrees with brute force."
\end{lstlisting}
At most three failed-probe lines are listed. $Z_{\text{brute}}$ itself is never disclosed. This limits output verbosity and information leakage.
 
\end{document}